\documentclass{article}
\usepackage{spconf,amsmath,graphicx}
\usepackage{amssymb,amsmath}
\usepackage{color}
\usepackage{mathrsfs}
\usepackage{multirow}
\usepackage{multicol}
\DeclareMathOperator*{\argmin}{arg\,min}
\DeclareMathOperator*{\argmax}{arg\,max}
\newcommand{\vect}[1]{\boldsymbol{#1}}
\DeclareMathOperator\vx{\boldsymbol{{x}}}
\DeclareMathOperator\vy{\boldsymbol{{y}}}
\DeclareMathOperator\ve{\boldsymbol{{\epsilon}}}
\DeclareMathOperator\vA{\boldsymbol{{A}}}
\DeclareMathOperator\vAtilde{\boldsymbol{{\tilde{A}}}}
\DeclareMathOperator\vAt{\boldsymbol{{\tilde{A}}}}
\DeclareMathOperator\vv{\boldsymbol{{v}}}
\DeclareMathOperator\vg{\boldsymbol{{\gamma}}}
\DeclareMathOperator\vgtilde{\boldsymbol{{\tilde{\gamma}}}}
\DeclareMathOperator\vd{\boldsymbol{{\delta}}}
\DeclareMathOperator\vmu{\boldsymbol{{\mu}}}
\DeclareMathOperator\vS{\boldsymbol{{\Sigma}}}

\DeclareMathOperator\vX{\boldsymbol{{X}}}
\DeclareMathOperator\vXt{\tilde{\boldsymbol{{X}}}}
\DeclareMathOperator\vxt{\tilde{\boldsymbol{{x}}}}
\DeclareMathOperator\vY{\boldsymbol{{Y}}}
\DeclareMathOperator\vE{\boldsymbol{{E}}}

\DeclareMathOperator\vV{\boldsymbol{{V}}}
\DeclareMathOperator\vI{\boldsymbol{{I}}}
\DeclareMathOperator\vDel{\boldsymbol{{\Delta}}}
\DeclareMathOperator\vP{\boldsymbol{{\Psi}}}

\newcommand{\imscale}{0.25}

\usepackage[nodisplayskipstretch]{setspace}
\DeclareMathOperator*{\inv}{^{-1}}
\usepackage{multirow}
\usepackage[font=small,skip=0pt]{caption}
\usepackage{subcaption}
\usepackage{bm}

\usepackage{tikz}
\usetikzlibrary{shapes}
\usetikzlibrary{positioning,arrows}
\usepackage[utf8x]{inputenc}
\usepackage{scalefnt}

\title{Robust Bayesian Method for Simultaneous Block Sparse Signal Recovery with Applications to Face Recognition}

 \name{{Igor Fedorov \sthanks{Igor Fedorov was partially supported by the San Diego Chapter of the ARCS Foundation, Inc}, Ritwik Giri, Bhaskar D. Rao and Truong Q. Nguyen}}
\address{Department of Electrical and Computer Engineering \\
University of California, San Diego\\
}

\begin{document}
\ninept
\maketitle
\begin{abstract}
In this paper, we present a novel Bayesian approach to recover simultaneously block sparse signals in the presence of outliers. The key advantage of our proposed method is the ability to handle non-stationary outliers, i.e. outliers which have time varying support. We validate our approach with empirical results showing the superiority of the proposed method over competing approaches in synthetic data experiments as well as the multiple measurement face recognition problem.
\end{abstract}


\section{Introduction}
Sparse Signal recovery (SSR) refers to algorithms which seek sparse solutions to underdetermined systems of equations \cite{eladbook}, which occur naturally when one seeks a representation of a given signal under an overcomplete dictionary. Overcomplete dictionaries have gained popularity in a wide range of applications because they are much more flexible than their undercomplete counterparts and lead to unique solutions under certain constraints, when sparsity has been enforced \cite{donoho2006stable}. Constraining the solution of underdetermined problems to be sparse represents prior knowledge about the solution and makes finding it tractable. In certain applications, structured sparsity, such as block sparsity, has been enforced on the desired coefficient vector, i.e., a small number of blocks of the solution are non-zero \cite{eldar2010block}.

SSR has become a very active research area in recent times because of its wide range of engineering applications. For example, in several popular computer vision problems, such as face recognition \cite{wright2009robust}, motion segmentation \cite{elhamifar2009sparse}, and activity recognition \cite{yang2008distributed}, signals lie in low-dimensional subspaces of a high dimensional ambient
space. An important class of methods to deal with this depends on exploiting  the notion of sparsity. Following this path, Sparse Representation based Classification (SRC) \cite{wright2009robust} was proposed and produced state of the art results in a face recognition (FR) task.

In many applications, we often encounter outliers in measurements, which leads traditional SSR algorithms to fail and necessitates the development of an outlier robust SSR algorithm. The need for outlier resistant SSR algorithms motivates our present work, in which we develop a robust SSR algorithm and extend it to recover simultaneous block sparse signals. To show the efficacy of our approach, we focus on FR, which refers to identifying a subject's face given a labeled database of faces. The pioneering work of Wright et al. \cite{wright2009robust} on SRC showed that a face classifier can be devised by using the downsampled images from the training database as a dictionary and considering the sparse representation of a given image under that dictionary as the "identity" of the person. In this scenario, it is intuitive to assume that the dictionary is broken up into blocks corresponding to each specific person and constraining the encoding of the image to be block-sparse leads to performance gains \cite{elhamifar2011robust}.

One significant challenge in the FR problem is dealing with occlusions. Occlusions are outliers within the SRC model because it assumes that the dictionary spans the space of all possible observations. A popular way to incorporate robustness to outliers into the SSR model is to assume that the outliers themselves have a sparse representation
 \cite{wright2009robust}, which has been shown to yield improved resilience to various forms of face occlusion and corruption \cite{wright2009robust, yang2011robust, wagner2012toward, elhamifar2011robust, li2013robust}. In certain cases, such as when the entire face is occluded or lighting conditions are extremely poor, FR within the SRC framework can yield unsatisfactory results because it is difficult to solve the single measurement vector (SMV) SSR problem. When possible, it is advantageous to acquire multiple measurements of the same source and instead solve the multiple measurement vector (MMV) problem. The MMV problem assumes that the support of the non-zero coefficients that encode each measurement does not change, while the actual values of the coefficients can vary. It is well known that the MMV problem yields much better recovery results than the SMV problem \cite{cotter2005sparse}.

In this work, we extend the SRC framework to the MMV case and consider performing FR when multiple images of the same subject, corrupted by non-stationary occlusions, are presented to the classifier. Our work is motivated, in part, by the person re-identification problem \cite{hirzer2011person, lisanti2015person, karanam2015sparse}. Srikrishna et al. \cite{karanam2015sparse} addressed the re-identification problem by applying SSR to each individual image of the subject and aggregating the results to form a global classifier. As such, \cite{karanam2015sparse} did not address the MMV nature of the problem. The main motivation behind our work is to enforce the prior knowledge that the input images correspond to the same person within the SSR process, while still maintaining resilience to time-varying occlusions.

Our SSR framework builds upon the hierarchical Bayesian framework discussed in \cite{tipping2001sparse,wipf2007empirical,zhang2011sparse}, known as Sparse Bayesian Learning (SBL).
 This choice is motivated by the superior recovery results obtained for the standard SSR problem \cite{wipf2007empirical, giri2015type} and the Bayesian framework is convenient for extensions to problems with structure \cite{zhang2013extension}. In this work, we extend the SBL framework to the MMV block-sparse case and explicitly model time-varying occlusions, referring to our method as robust SBL (Ro-SBL).

\subsection{Contributions}
\begin{itemize}
\item We introduce a novel hierarchical Bayesian Robust SSR algorithm, Ro-SBL, for solving the MMV block-sparse problem with time-varying outliers. This work has connections to \cite{li2013robust}, where a Robust Block Sparse Bayesian Learning (BSBL) method was proposed. In contrast with our work, BSBL only considered the SMV problem and did not harness the ability of the SBL framework to capture non-stationary outliers.


\item We validate our proposed method with synthetic data results and also apply our method to a robust simultaneous FR task. Unlike \cite{karanam2015sparse}, our proposed approach exploits the prior knowledge that the input images correspond to the same person within the SSR process.
\end{itemize}
\vspace{-1.5em}

\section{Ro-SBL for Simultaneous Block Sparse Recovery}
The signal model for simultaneous block sparse recovery is given by
\begin{align}\label{eq:model}
\vY = \vA \vX + \vE + \mathbf{V}
\end{align}
where, $\vY \in \mathbb{R}^{n \times L}$ is the matrix of $L$  measurements,  $\vect{A} \in \mathbb{R}^{n \times m}$ is the dictionary, $\mathbf{V} \in \mathbb{R}^{n \times L}$ is the independent and identically distributed (IID) Gaussian noise term with mean zero and variance $\sigma^2$, $\vX \in \mathbb{R}^{m \times L} $ is the encoding of the measurements under $\vA$, and $\vE \in \mathbb{R}^{n \times L}$ is the matrix containing the outliers in the measurements.

The key assumption in the MMV problem is that, if a given column of $\vect{A}$ is activated (i.e. its corresponding coefficient in $\vX$ is non-zero) for one of the measurements, then it will be activated for all of the measurements \cite{cotter2005sparse}. This means that the same set of basis vectors have been used to generate all of the measurements, which is reflected in the encoding matrix $\vX$ in the form of joint sparsity, i.e. $\lbrace \vx_{(:,i)} \rbrace_{i=1}^L$ share the same support, where $\vx_{(:,i)}$ is the $i$'th column of $\vX$ \cite{cotter2005sparse}. Within a Bayesian framework, the joint sparsity assumption translates to placing a prior on the \textit{rows } of $\vX$. In the context of our work, we build upon the extension of the SBL framework to the MMV problem in \cite{wipf2007empirical} and adopt a hierarchical prior, namely a Gaussian Scale Mixture (GSM), over the rows of $\vX$:
\begin{equation}
p\left(\mathbf{x}_{(j,:)} |  \gamma_j \right) =  \mathsf{N} \left(\mathbf{x}_{(j,:)};0, \gamma_j \vI_L \right)
\end{equation}
where, $\mathbf{x}_{(j,:)}$ denotes the $j$'th row of $\vX$ and $\gamma_j$ is the unknown variance hyperparameter. In addition, we also consider block sparsity in each $\vx_{(:,i)}$, where the block structure is shared among all of the encoding vectors. Assuming that the support of $\vx_{(:,i)}$ is separated into disjoint sets $\mathscr{G}_g, 1 \leq g \leq G$, which are known a-priori and shared across all $i$, $p(\vX)$ is amended to reflect that each of the rows in a given group $\mathscr{G}_g$ share the same $\vg_g$ \cite{zhang2013extension}:
\begin{equation}
p(\vX) = \prod_{g=1}^G \prod_{j \in \mathscr{G}_g} p\left(\mathbf{x}_{(j,:)} | \gamma_g\right)
\end{equation}

Although the joint block sparsity constraint is a valid one for the encoding matrix $\vX$, it does not hold for the outlier matrix $\vE$ since the outliers could be non-stationary, i.e., time varying. Therefore, we will treat each $\ve_{(:,i)}$ independently and not constrain the outliers to share the same support across all measurements. As such, we adopt a sparsity enforcing GSM prior on $\ve_{(:,i)}$, which induces the following prior on $\vE$:
\begin{equation}
p (\vE) = \prod_{j=1}^n \prod_{i=1}^L p(\epsilon_{ji}| \delta_{ji}) = \prod_{j=1}^n \prod_{i=1}^L \mathsf{N}(\epsilon_{ji}; 0, \delta_{ji})
\end{equation}
This set of assumptions is unique to this work and is motivated by the FR task.
\vspace{-1em}
\subsection{Incorporating Robustness to Outliers}
\vspace{-0.2em}
For an SMV problem, \cite{li2013robust}\cite{jin2010algorithms} showed that sparse (under the standard basis) outliers can be incorporated into the well known SBL framework by introducing a simple modification to the dictionary $\vA$. In the present work, we extend this idea to the MMV case, which results in the following modification to the signal model in \eqref{eq:model}:
\begin{align}\label{eq:zhilin}
\vY = \underbrace{\begin{bmatrix}
\vA & \vI_n
\end{bmatrix}}_{\boldsymbol{\tilde{A}}} \underbrace{\begin{bmatrix}
\vX \\ \vE
\end{bmatrix}}_{\vXt} + \vV.
\end{align}
Note that \eqref{eq:zhilin} and \eqref{eq:model} are equivalent, but, as will be shown next, the signal model in \eqref{eq:zhilin} lends itself much more nicely to a closed form inference procedure.
\vspace{-1em}
\subsection{Ro-SBL Inference Procedure}
The goal of the inference procedure is to estimate the hyperparameters $\vg = \begin{bmatrix}
\gamma_1 & \cdots & \gamma_G
\end{bmatrix}^T$ and $\vDel = \begin{bmatrix}
\vd_{(:,1)} & \cdots & \vd_{(:,L)}
\end{bmatrix}$, where $\vd_{(:,i)} = \begin{bmatrix}
\delta_{1i} &\cdots& \delta_{ni}
\end{bmatrix}^T$. As in \cite{tipping2001sparse}\cite{wipf2007empirical}, we adopt an Expectation Maximization (EM) procedure where we treat $\vXt$ as the hidden data. In the E-step, we seek the expectation of the complete data $(\vY,\vXt,\vg,\vDel)$ log likelihood under the posterior $p(\vXt | \vY, \vg^t,\vDel^t,\sigma^2)$, where $t$ denotes the iteration index. Because $\lbrace \vxt_{(:,i)} \rbrace_{i=1}^L$ are conditionally independent given $\vY$, $\vg$, and $\vDel$, the E-step reduces to
\begin{align}\label{eq:Q}
\begin{split}
& Q(\vg,\vDel, \sigma^2, \vg^t,\vDel^t) = \\
& \sum_{i=1}^L E_{\vxt_{(:,i)} | \vy_{(:,i)},\vg^t,\vd_{(:,i)},\sigma^2} \left[ \log p\left(\vy_{(:,i)},\vxt_{(:,i)},\vg,\vd_{(:,i)},\sigma^2 \right)\right].
\end{split}
\end{align}
The posterior needed to compute \eqref{eq:Q} is given by $\mathsf{N}\left( \vxt_{(:,i)} ; \vmu^i, \vS^i \right)$, where $\vmu^i$ and $\vS^i$ are given by \cite{tipping2001sparse}\cite{wipf2007empirical}
\begin{align*}
\begin{split}
\vmu^i &= \vP^i \vAtilde^T \left(\sigma^2 \boldsymbol{I}_{m+n} + \vAtilde \vP^i \vAtilde^T\right)\inv \vy_{(:,i)} \\
\vS^i &= \vP^i - \vP^i \vAtilde^T \left(\sigma^2 \boldsymbol{I}_{m+n}  + \vAtilde \vP^i \vAtilde^T \right)\inv \vAtilde \vP^i
\end{split}
\end{align*}
where $\vP^i$ is a diagonal matrix containing $\begin{bmatrix}
\vgtilde^T & \vd_{(:,i)}^T
\end{bmatrix}$ on the diagonal and $\vgtilde \in \mathbb{R}^m$ is set to $\tilde{\gamma}_j = \gamma_g$ for $j \in \mathscr{G}_g$. Unlike \cite{wipf2007empirical}, where the covariance of the posterior is shared for all $i$, the covariance is a function of $i$ here because each $\vxt_{(:,i)}$ consists of $\vx_{(:,i)}$, whose support does not vary with $i$, and $\ve_{(:,i)}$, whose support does vary.

In the M-step, $Q(\vg,\vDel,\sigma^2, \vg^t,\vDel^t)$ is maximized with respect to $(\vg,\vDel, \sigma^2)$, leading to the update rules:
\begin{align}\label{eq:update rules mmv}
\begin{split}
&\gamma_g = \sum_{i=1}^L\sum_{j \in \mathscr{G}_g} \frac{\vS_{jj}^i + \left(\vmu_j^i\right)^2}{\vert \mathscr{G}_g \vert L}  \\
&\delta_{ji} = \vS_{\tilde{j} \tilde{j}}^i + \left(\vmu_{\tilde{j}}^i\right)^2 \; ,\; 1 \leq j \leq n \; , \; \tilde{j} = j + m \\
&\sigma^2 = \\
&\sum_{i=1}^L \frac{\left\Vert \vy_{(:,i)} \right\Vert^2 -2\vy_{(:,i)}^T \vAt \vmu^i + \mathsf{tr}\left( \vAt^T \vAt \left( \vS^i + \vmu^i \left(\vmu^i\right)^T\right)\right)  }{L n}
\end{split}
\end{align}
where $\mathsf{tr}(\cdot)$ refers to the trace operator.

Upon convergence of the EM algorithm to the estimates $\hat{\vg}$ and $\hat{\vDel}$, $\vX$ and $\vE$ can be estimated using the maximum a-posteriori (MAP) estimator:
\begin{align*}
\begin{bmatrix}
\hat{\vx}_{(:,i)}^T & \hat{\ve}_{(:,i)}^T
\end{bmatrix}^T  = \argmax_{\vxt_{(:,i)}} p\left( \vxt_{(:,i)} | \vy_{(:,i)},\hat{\vg},\hat{\vd}_{(:,i)},\sigma^2 \right) = \hat{\vmu}^i
\end{align*}
\vspace{-1.5em}
\section{Results}
\vspace{-0.5em}
\subsection{Synthetic Data Results}
To validate the proposed method, we conducted SSR experiments on synthetic data. To generate the synthetic data, we begin by randomly selecting $s$ sets from $\lbrace \mathscr{G}_g \rbrace_{g=1}^G$ and generate $\vx_{(:,i)}$ such that the non-zero elements are indexed by one of the selected sets. We use equally sized blocks of length $8$. The non-zero elements of $\vx_{(:,i)}$ are drawn from the $\mathsf{N}(0,1)$ distribution.  We generate $\vA \in \mathbb{R}^{80 \times 160}$ by drawing its elements from the $\mathsf{N}(0,1)$ distribution and normalizing the columns to have unit $\ell_2$ norm. Finally, we use the robust modeling strategy and replace
 $\vA$ by  $\tilde{\vA} = \begin{bmatrix}
\vA & \vI_n \end{bmatrix}$. In order to simulate a noisy SSR scenario, we generate $\vv_{(:,i)}$ by drawing its elements from the $\mathsf{N}(0,1)$ distribution and $\ve_{(:,i)}$ by drawing its elements form the student-t distribution with one degree of freedom. Finally, we generate observations $\vy_{(:,i)}$ according to \eqref{eq:model} after scaling $\vv_{(:,i)}$ and $\ve_{(:,i)}$ to achieve a specified Signal-to-Gaussian noise ratio (SGNR) and Signal-to-Outlier noise ratio (SONR).

Let $\hat{\vX}$ denote the approximation to $\vX$ generated by the SSR algorithm. We measure the quality of the recovery using the relative $\ell_2$ error: $\frac{1}{L}\sum_{i=1}^L \frac{\Vert \vx_{(:,i)} - \boldsymbol{\hat{x}}_{(:,i)} \Vert^2}{\Vert \vx_{(:,i)} \Vert^2}$. We performed the synthetic data experiment 500 times and report the average performance results.

We compare the performance of the proposed method to several standard SSR algorithms. As a baseline, we use the $\ell_1$ SSR approach and the block sparse extension of the $\ell_1$ approach, the $\ell_2-\ell_1$ block SSR algorithm (also known as Group LASSO \cite{bakin1999adaptive}), which seeks
\begin{align}\label{eq:l2-l1}
\boldsymbol{\hat{x}} &= \argmin_{\vx} \Vert \vA \vx - \vy\Vert_2^2 + \lambda \sum_{g=1}^G \Vert \vx_{\mathscr{G}_g} \Vert_2.
\end{align}
Note that \eqref{eq:l2-l1} reduces to the $\ell_1$ SSR objective function when each element of $\vx$ is a separate group. We use the SLEP \cite{liu2009slep} software package to solve the $\ell_1$ and $\ell_2-\ell_1$ problems. For comparison purposes we naively extend the $\ell_1$ and $\ell_2-\ell_1$ approaches to the MMV case by solving each MMV problem as $L$ independent SMV problems.

We also compare our approach with Block-SBL (BSBL) \cite{zhang2013extension, li2013robust}, which is a hierarchical Bayesian framework for solving the SMV block-sparse recovery problem.  We naively extend BSBL to the MMV case by assuming that the outliers have stationary support, denoting the resulting algorithm as M-BSBL. In the context of the signal model in \eqref{eq:model}, M-BSBL corresponds to assuming \textit{row sparsity} on $\vE$.

\begin{figure}
\centering
\begin{subfigure}[b]{0.45\textwidth}
\centering\includegraphics[height=15em, width = \linewidth]{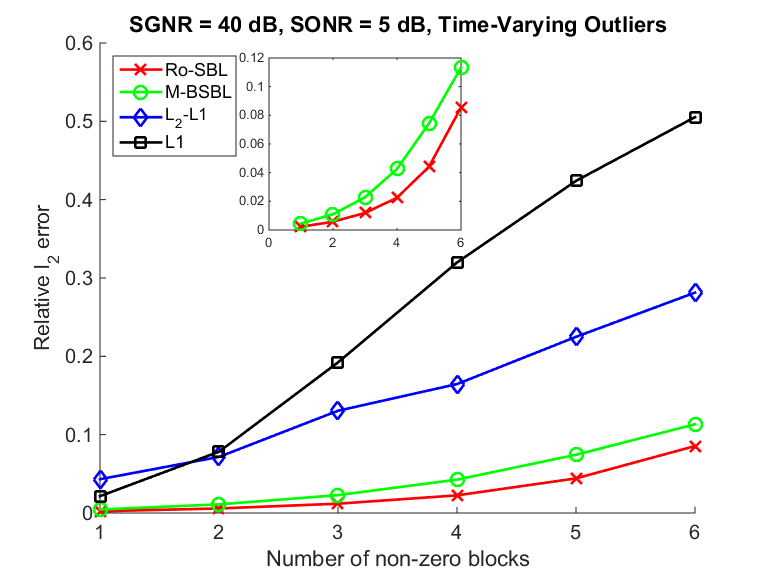}
\caption{Time-Varying outliers}
\label{fig:tvo}
\end{subfigure}

\begin{subfigure}[b]{0.45\textwidth}
\centering\includegraphics[height=15em, width = \linewidth]{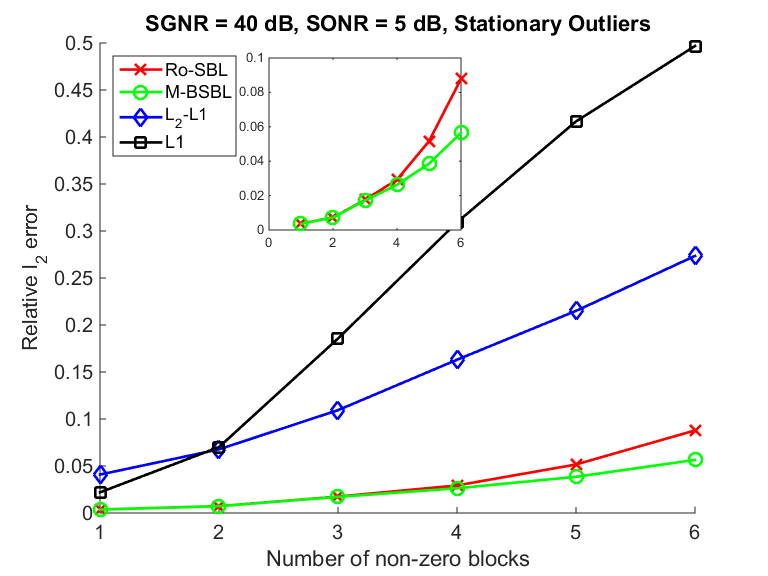}
\caption{Stationary outliers}
\label{fig:so}
\end{subfigure}

\caption{Comparison of SSR algorithms on synthetic data}
\vspace{-0.5em}
\end{figure}

Simulation results for a 5 measurement SSR problem with 40 dB SGNR and 5 dB SONR are shown in Fig. \ref{fig:tvo}. M-BSBL and Ro-SBL drastically outperform the $\ell_1$ and $\ell_2-\ell_1$ SSR approaches, which shows that hierachical Bayesian approaches outperform deterministic methods even in the challenging SSR setup considered. In addition, since the Bayesian approaches explicitly model the MMV nature of the problem, this result shows that significant improvements can be achieved by incorporating the prior knowledge that the support of $\vx_{(:,i)}$ does not change with $i$ in the SSR algorithm. We observe a $25\%-51\%$ improvement in relative $\ell_2$ error from Ro-SBL compared to M-BSBL for $s \leq 6$. This suggests that Ro-SBL is better able to capture outliers due to its superior model.

For illustrative purposes, we conducted the same synthetic data experiment, with the exception that the outliers were constrained to be the same for all $i$. The results are shown in Fig. \ref{fig:so}. As expected, the performance gains of the hierachical Bayesian approaches over the deterministic approaches carries over into the stationary outlier scenario. It is important to note that M-BSBL slightly outperforms Ro-SBL because, in this case, the M-BSBL signal model is better fitted to estimate the outliers since M-BSBL assumes that the outliers are stationary and enjoys the advantages of MMV modeling, even for outliers.


\begin{table*}[t]
	\centering
	\begin{tabular}{|c|c|c|c|c|c|c|c|c|c|c|}
	\hline
		   &	\multicolumn{2}{c|}{$10\%$}	&	\multicolumn{2}{c|}{$20\%$}	&	\multicolumn{2}{c|}{$30\%$}	&	\multicolumn{2}{c|}{$40\%$}	& \multicolumn{2}{c|}{$50\%$}   \\ \hline
		   & $L = 1$ & $L = 5$ &$L = 1$ &$L = 5$ &$L = 1$ &$L = 5$ &$L = 1$ &$L = 5$ &$L = 1$ &$L = 5$ \\ \hline
		SRC \cite{wright2009robust}	&   89.72 & 100    &	83.61 & 100 & 71.29 &	97.81   & 54.24 &	92.54 	& 35.97 & 67.98   \\ \hline
		$P_{\ell_2-\ell_1}$ \cite{elhamifar2011robust}	&	91.60 & 100& 84.50 & 100& 69.90&	96.93&54.08 &	93.42 & 39.07 & 72.37\\ \hline
		M-BSBL \cite{li2013robust}	&	--- & 100	&	--- & 100 & --- &	99.12&	---& 94.30 & --- & 76.75\\ \hline
		Ro-SBL (proposed)	&	\textbf{91.92} & \textbf{100} &	\textbf{91.11} & \textbf{100}& \textbf{83.52}&	\textbf{100}&	\textbf{65.58}& \textbf{100} & \textbf{41.44} & \textbf{91.22} \\ \hline
	\end{tabular}
	
	\vspace{1em}
	\caption{Face classification accuracy results for $L = 1$ and $L = 5$ for various occlusion rates}
	\label{table:classification}
\end{table*}

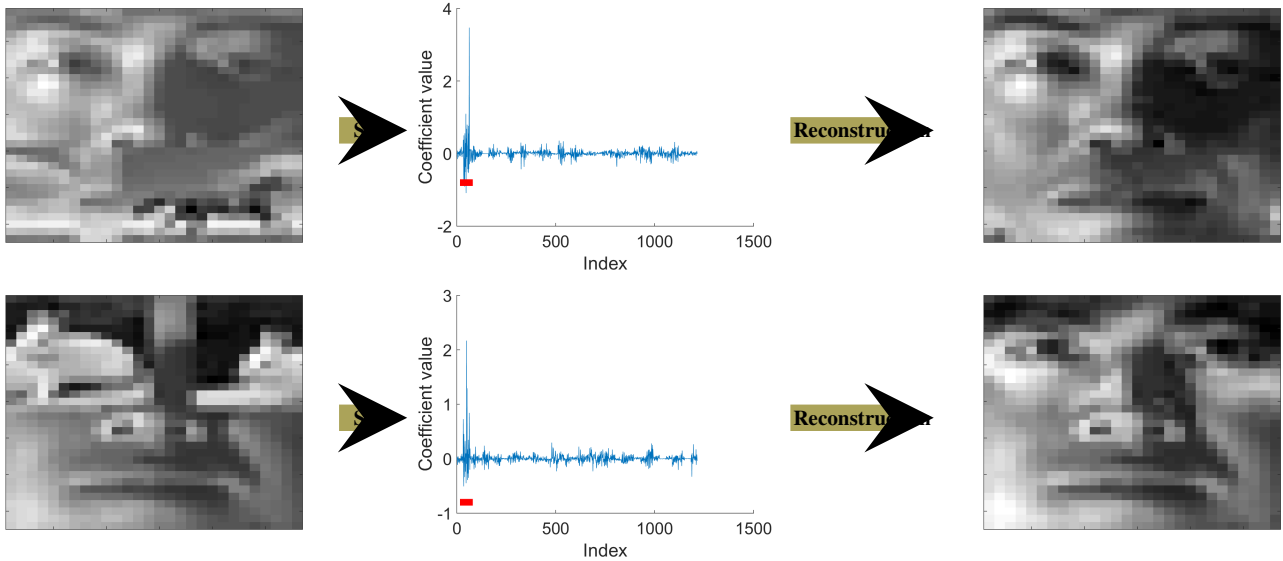
\begin{figure*}
	\centering
	\begin{multicols}{2}
	\centering
\begin{tikzpicture}
    \node[anchor=north west,inner sep=0] at (0,0) (c1) {\includegraphics[scale=\imscale]{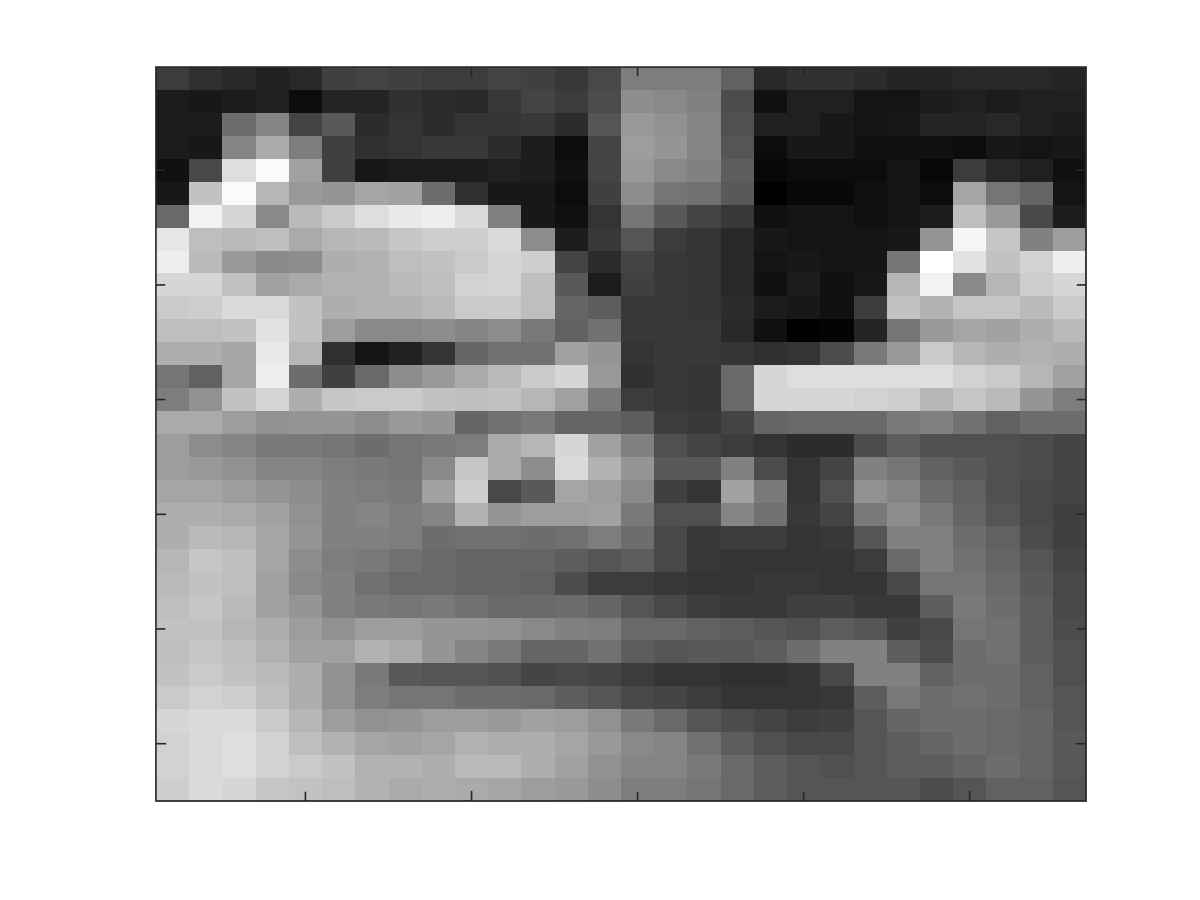}};
        \node[anchor=north west,inner sep=0] at (6,0) (c2) {\includegraphics[scale=\imscale]{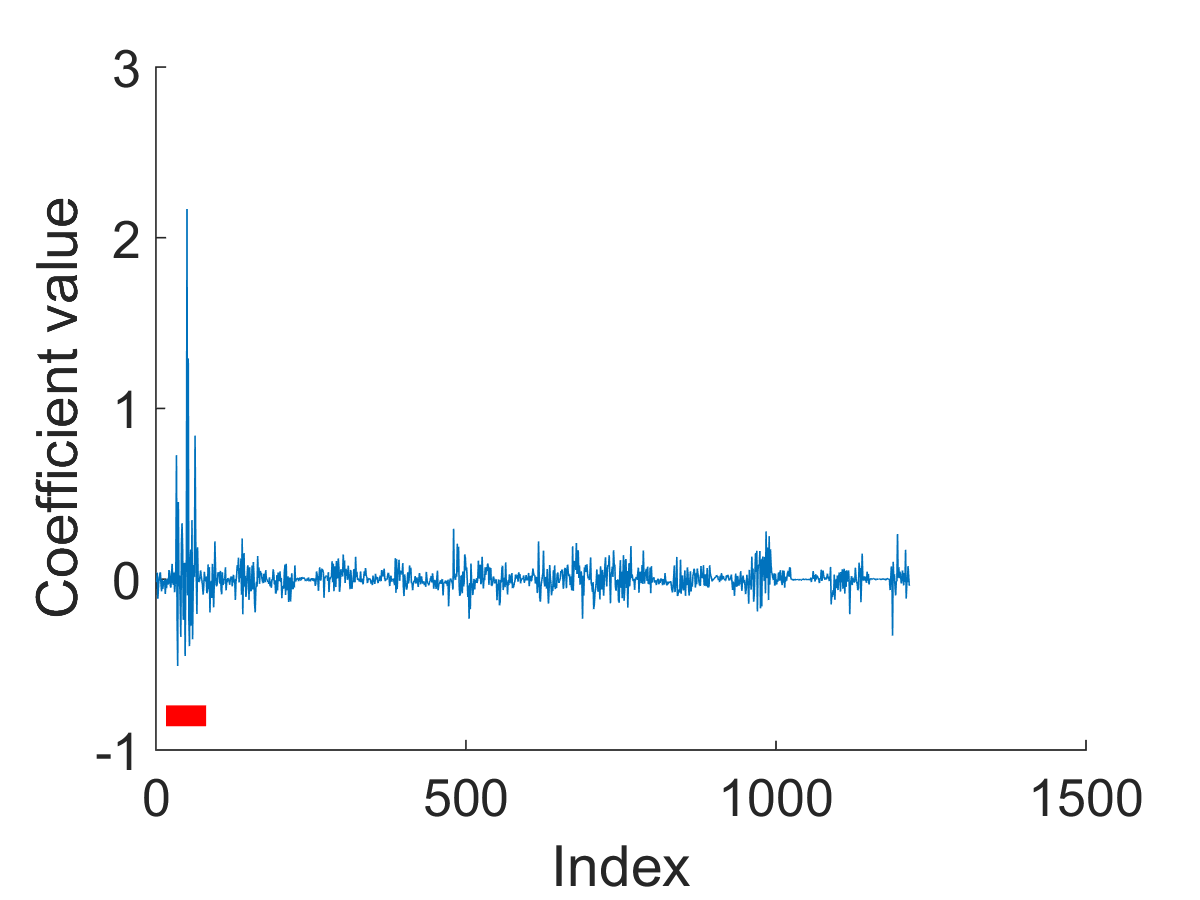}};
                \node[anchor=north west,inner sep=0] at (13,0) (c3) {\includegraphics[scale=\imscale]{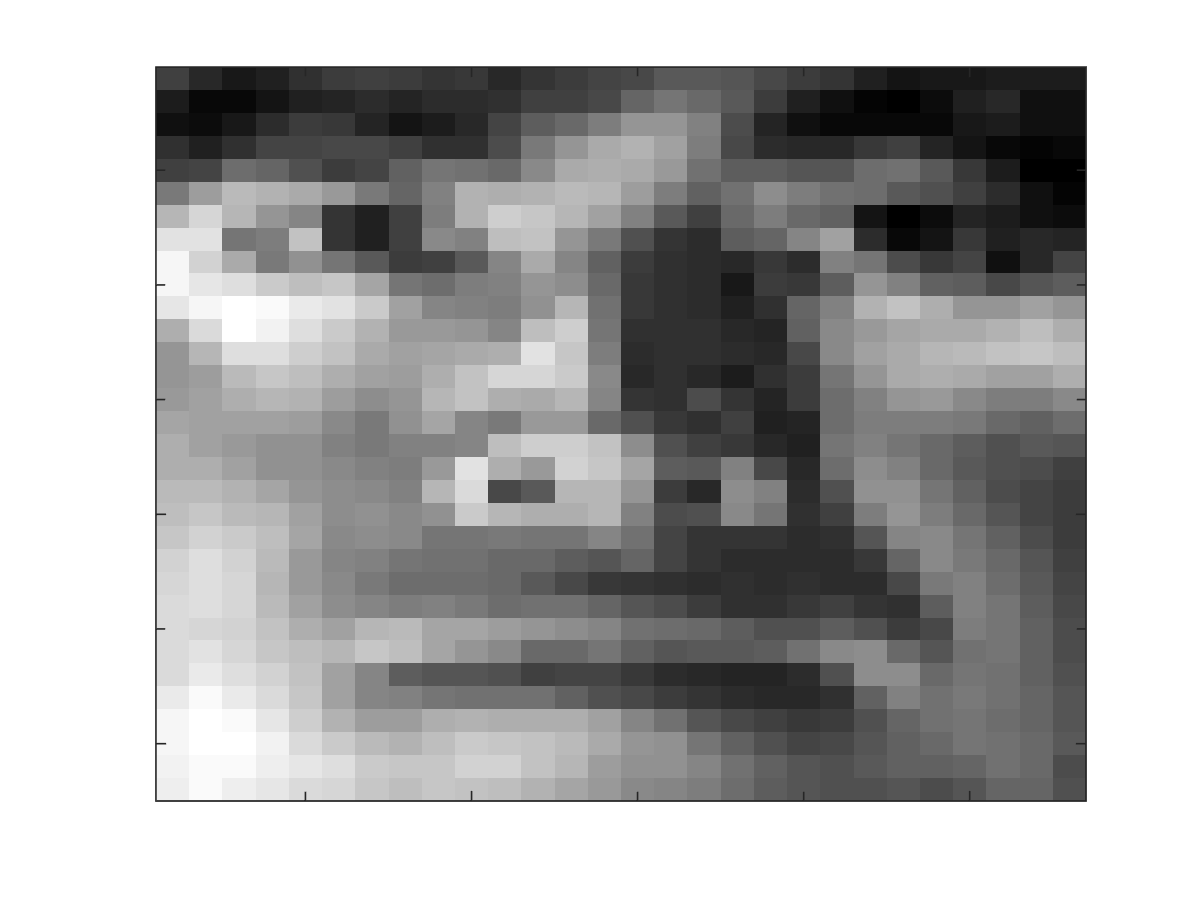}};
    \node[anchor=south west,inner sep=0] at (0,0) (c4) {\includegraphics[scale=\imscale]{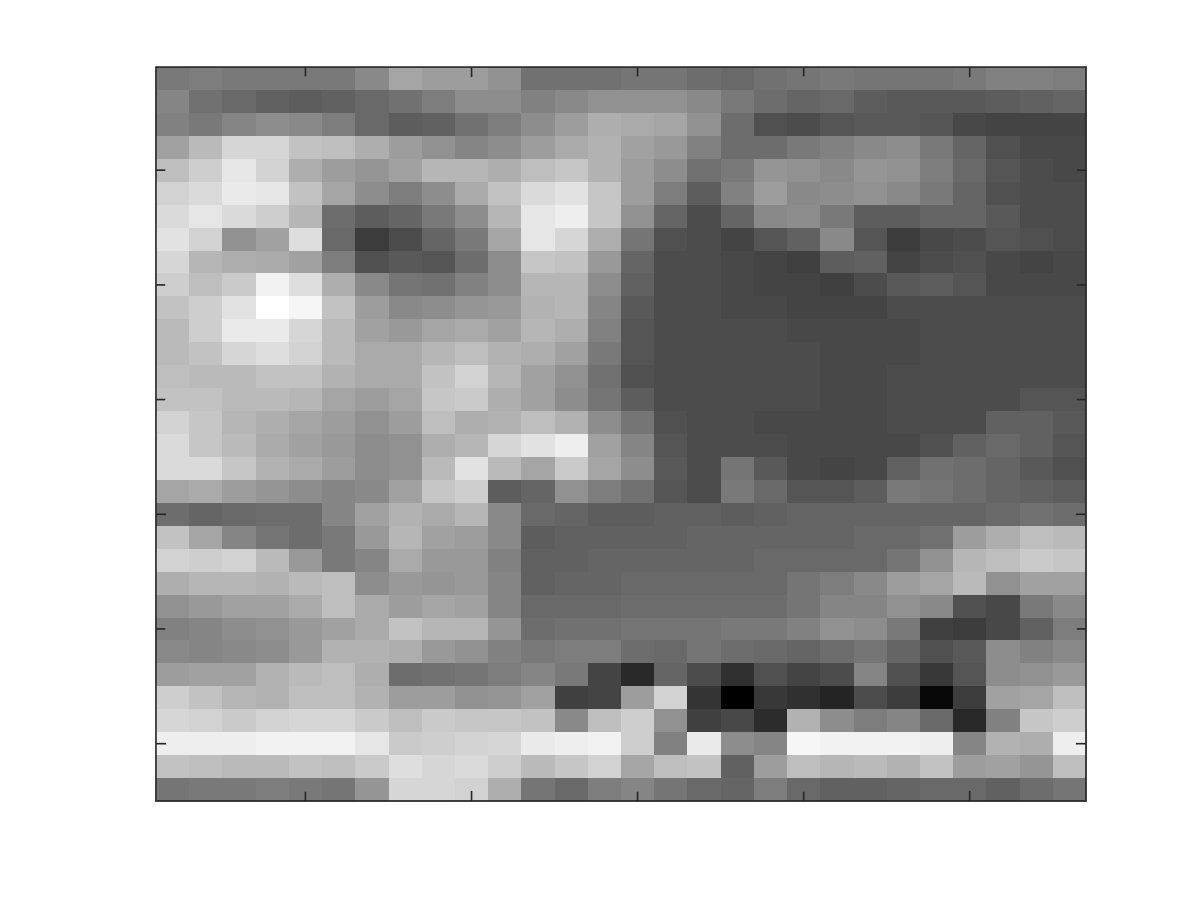}};
        \node[anchor=south west,inner sep=0] at (6,0) (c5) {\includegraphics[scale=\imscale]{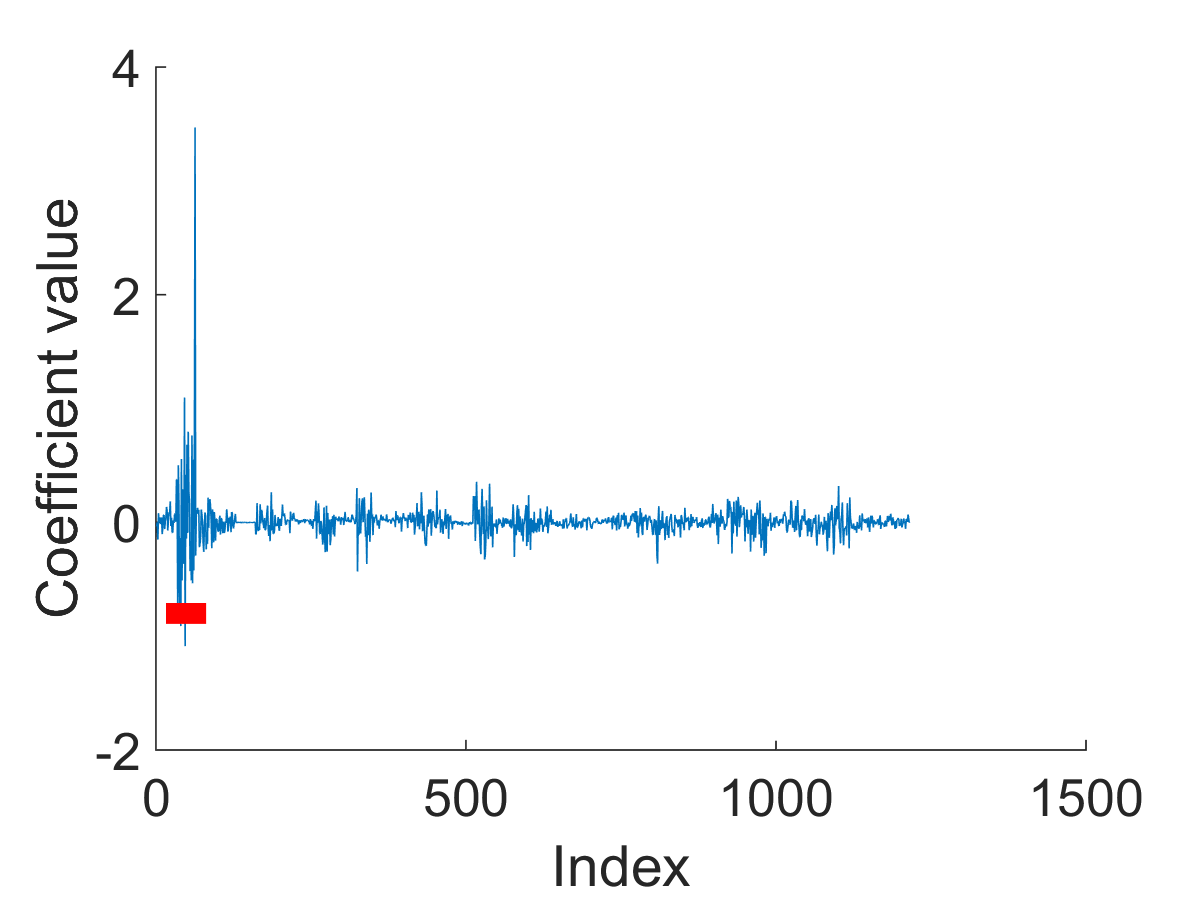}};
                \node[anchor=south west,inner sep=0] at (13,0) (c6) {\includegraphics[scale=\imscale]{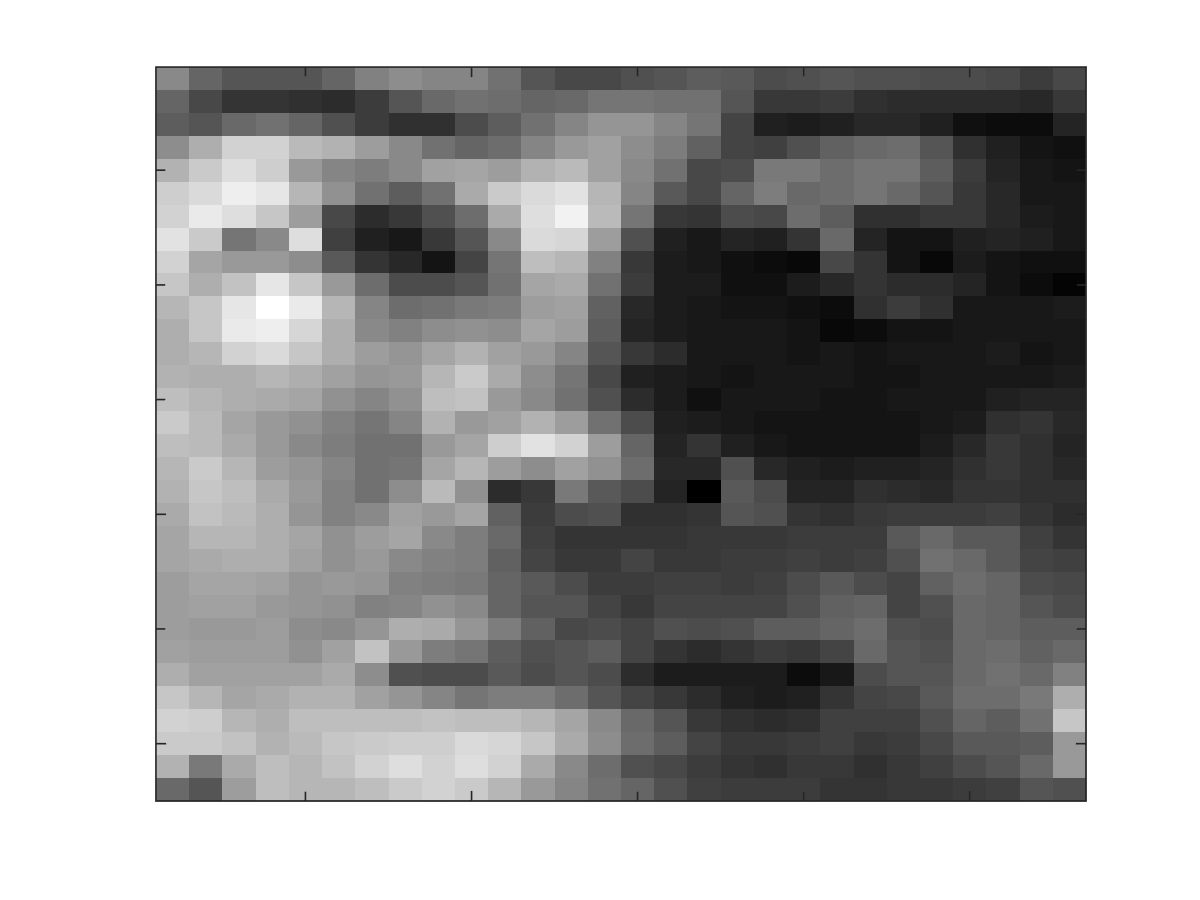}};

                \draw[draw=yellow!60!black,line width=10pt,-stealth] (c1)  -- (c2)
        node[midway,text=black,font=\footnotesize\bfseries]{SSR};

                        \draw[draw=yellow!60!black,line width=10pt,-stealth] (c2)  -- (c3)
        node[midway,text=black,font=\footnotesize\bfseries]{Reconstruction};

                        \draw[draw=yellow!60!black,line width=10pt,-stealth] (c4)  -- (c5)
        node[midway,text=black,font=\footnotesize\bfseries]{SSR};

                                \draw[draw=yellow!60!black,line width=10pt,-stealth] (c5)  -- (c6)
        node[midway,text=black,font=\footnotesize\bfseries]{Reconstruction};
\end{tikzpicture}
	\end{multicols}
	\caption{Example of face reconstruction using Ro-SBL. The red bar denotes coefficients corresponding to the true subject class.}
	\label{fig:reconstruction}
\end{figure*}
\vspace{-1em}
\subsection{FR Results}
\vspace{-0.2em}
In this section, we present results demonstrating the efficacy of the proposed method in a FR task. We use the Extended Yale B Database \cite{georghiades2001few}, which consists of 2441 images of 38 subjects under various illumination conditions. Each $192 \times 168$ image is a frontal perspective of the subject's face and has been cropped such that only the face can be seen. We randomly split the database into training and test sets. Following the SRC framework, we downsample the images in the training set by $\frac{1}{12}$, vectorize the result, and concatenate the vectors to form the dictionary $\vA$:
\begin{align*}
\vA = \begin{bmatrix}
\underbrace{\cdots \vA^1 \cdots}_{Block 1} | & \underbrace{\cdots \vA^2 \cdots}_{Block 2} |& \cdots \cdots |& \underbrace{\vA^{38}}_{Block 38}
\end{bmatrix}
\end{align*}
where $\vA^k$ consists of training images from the $k$'th subject. Note that this automatically introduces a block structure in $\vA$. As before, we replace $\vA$ by $\boldsymbol{\tilde{A}}$.

In the testing phase, we seek to identify a given subject from $L$ images of that subject's face. To simulate time-varying occlusions, we occlude each image by one of 10 animal images, choosing the location of the occluding image randomly. Given $L$ observations, we use one of the SSR algorithms to estimate $\hat{\vX}$ and $\hat{\vE}$. Similar to the person re-identification classifier presented in \cite{karanam2015sparse}, we label the test images using
\begin{align*}
k^* = \argmin_k \sum_{i = 1}^L \left\Vert \vy_{(:,i)} - \boldsymbol{{A}} \left(\boldsymbol{\phi}^k \odot \boldsymbol{\hat{x}}_{(:,i)}\right) - \boldsymbol{\hat{\epsilon}}_{(:,i)}\right\Vert^2
\end{align*}
where $\boldsymbol{\phi_{\mathscr{G}_k}}^k = 1$ and $\boldsymbol{\phi_{\mathscr{G}_c}}^k = 0, \forall c \neq k$.

For each test subject, around $30$ test images were available and the classification experiment was run $\left \lfloor{\frac{30}{L}}\right \rfloor$ times. We report averaged classification results in Table \ref{table:classification} for two cases: $L=1$, i.e. an SMV FR problem considered in \cite{yang2011robust, wright2009robust, elhamifar2011robust}, and $L=5$,  i.e. an MMV problem considered in person re-identification \cite{karanam2015sparse}. Note that our proposed algorithm becomes equivalent to M-BSBL for $L=1$, hence we do not report M-BSBL results for this case in Table \ref{table:classification}. It is evident from the results that, for every algorithm, the $L=5$ case leads to much better classification accuracy compared to $L=1$, which corroborates the well known result that MMV modeling is superior to SMV in harsh conditions.

In all cases, Ro-SBL performs better than all other competing algorithms. For low occlusion rates ($10\%$), all of the competing algorithms perform comparably,  despite the fact that only Ro-SBL explicitly models non-stationary occlusions. This result can be explained by the fact that $10\%$ occlusion represents a minor fraction of the overall image area and a significant amount of facial features remain un-occluded. On the other hand, Ro-SBL drastically outperforms the other algorithms at high occlusion rates ($50 \%$) with $L=5$, which can be attributed to the fact that a large portion of the facial features in each image are occluded and the SSR algorithm is forced to use all 5 measurements to jointly recover $\hat{\vX}$. Since Ro-SBL models outliers more accurately than the other SSR methods, it is better able to approximate the identity of the occluded subject.

Finally, as a visual example of how the proposed method performs face classification, we show that the occlusion can be removed from the test image by considering $\vA\vx_{(:,i)}$ as the estimate of the original, un-occluded test image. The face reconstruction result is shown in Fig. \ref{fig:reconstruction} (results are generated using $L = 5$ and a downsampling factor of $\frac{1}{6}$ was used for visualization purposes), which shows that the proposed method removes much of the occluding image and provides a relatively good reconstruction of the original face. Moreover, the coefficient plots show that the dominant coefficients all reside in the block corresponding to the test subject's index in $\vA$.
\vspace{-0.9em}
\section{Conclusion}
\vspace{-0.9em}
In this article, we have proposed a novel robust sparse recovery algorithm based on the well known SBL framework to recover simultaneous block sparse signals in presence of time varying outliers. Along with validating our method on synthetic data, we show the efficacy of our approach in simultaneous FR in the presence of time varying outliers.

\bibliography{ManuscriptBib}
\end{document}